\documentclass[10pt,journal,compsoc]{IEEEtran}

\usepackage{times}
\usepackage{epsfig}
\usepackage{graphicx}
\usepackage{amsmath}
\usepackage{amssymb}
\usepackage{multirow}
\usepackage{bbm}
\usepackage{comment}
\usepackage{float}
\usepackage{xcolor}
\usepackage{array}
\usepackage{blindtext}
\newcommand{\etal}{\textit{et al}.}
\newcommand{\ie}{\textit{i.e.}}
\newcommand{\eg}{\textit{e.g.}}

\ifCLASSOPTIONcompsoc
  \usepackage[nocompress]{cite}
\else
  \usepackage{cite}
\fi

\ifCLASSINFOpdf
\else
\fi

\begin{document}
%
\title{Instance Image Retrieval by Learning Purely From Within the Dataset}
\author{Zhongyan Zhang, Lei Wang,~\IEEEmembership{Senior Member,~IEEE}, Yang Wang, Luping Zhou,~\IEEEmembership{Senior Member,~IEEE}, Jianjia Zhang, Peng Wang, and Fang Chen}
\IEEEtitleabstractindextext{%
\begin{abstract}
Quality feature representation is key to instance image retrieval. To attain it, existing methods usually resort to a deep model pre-trained on benchmark datasets or even fine-tune the model with a task-dependent labelled auxiliary dataset. Although achieving promising results, this approach is restricted by two issues: 1) the domain gap between benchmark datasets and the dataset of a given retrieval task; 2) the required auxiliary dataset cannot be readily obtained. In light of this situation, this work looks into a different approach which has not been well investigated for instance image retrieval previously: {can we learn feature representation \textit{specific to} a given retrieval task in order to achieve excellent retrieval?} Our finding is encouraging. By adding an object proposal generator to generate image regions for self-supervised learning, the investigated approach can successfully learn feature representation specific to a given dataset for retrieval. This representation can be made even more effective by boosting it with image similarity information mined from the dataset. As experimentally validated, such a simple ``self-supervised learning + self-boosting'' approach can well compete with the relevant state-of-the-art retrieval methods. Ablation study is conducted to show the appealing properties of this approach and its limitation on generalisation across datasets. 
\end{abstract}

}

\maketitle

\IEEEdisplaynontitleabstractindextext

%
\IEEEpeerreviewmaketitle

\IEEEraisesectionheading{\section{Introduction}\label{sec:introduction}}
\IEEEPARstart{I}nstance image retrieval aims to find from a dataset the images containing the objects visually similar to a query object. A feature representation that faithfully reflects the visual similarity is crucial to this task. Significant progress has been made to pursue such a feature during the past two decades, from early global handcrafted features, through local invariant features, to recent deep learning based ones~\cite{zheng2017sift,babenko2014neural}.    

Instance image retrieval is unsupervised by nature, \ie, no label information is usually available to the images in the dataset for retrieval. As a result, existing retrieval methods often rely on external deep models to extract feature representation. The models are commonly pre-trained by generic large-scale image benchmarks such as ImageNet~\cite{deng2009imagenet}. The generality of deep features makes this approach convenient and able to produce reasonable performance for various retrieval tasks. Recently, several methods further enhance pre-trained deep models with auxiliary labelled datasets that contain images having a similar nature to those in a given retrieval dataset. This approach has shown the state-of-the-art retrieval performance~\cite{cao2020unifying,gordo2017end,DBLP:conf/eccv/NgBTM20,radenovic2018fine,revaud2019learning}.    

While enjoying the above merits, the existing approach also experiences the following two issues due to the reliance on pre-trained deep models or auxiliary datasets. First, domain gap could exist between the benchmark image dataset used to pre-train the deep model and the dataset of a given retrieval task. When this gap is not negligible, the efficacy of the pre-trained deep feature representation will reduce. Second, the requirement to access an auxiliary dataset that is labelled and shares a similar nature as the dataset for retrieval is hard to be generally met. When this turns out to be true, fine-tuning a pre-trained deep model for retrieval will become infeasible. 

The root of the above dilemma lies at the reliance on external models or datasets to learn feature representation for retrieval. This naturally leads to asking the following question: 
\textit{can we achieve promising retrieval by learning feature representation purely from within a given retrieval dataset, without leveraging any external deep models or auxiliary datasets?}         

An immediate concern on this approach may be the general applicability of the feature representation obtained in this way. After all, the representation learned purely from within a particular dataset may not generalise to other ones, which is in contrast to the merit of the existing approach which uses pre-trained deep models. Nevertheless, we argue that i) as aforementioned, the general applicability of pre-trained deep models does not always hold but could be adversely affected by domain gap; ii) if the approach of learning purely from within a dataset can achieve retrieval that is comparable to or even better than the state-of-the-art, it will have a clear advantage of not relying on any external deep models or auxiliary datasets; and iii) in many practical applications of image retrieval (such as in archives, museums, or medical repositories), the distribution of involved dataset does not change dynamically but usually has a static or relatively stable nature. In this case, compared with the property of general applicability, how to maximise retrieval performance will be a more important one when feature representation is considered. In light of these, this work believes that the question asked above has its value and a thorough investigation shall be conducted to answer it. 

To answer this question, this paper develops a novel framework built upon recent self-supervised learning techniques to learn feature representation for the task of instance image retrieval. Note that different from most existing works on self-supervised learning~\cite{chen2020simple,he2020momentum,DBLP:conf/icml/ZbontarJMLD21, DBLP:conf/nips/GrillSATRBDPGAP20}, this work does not aim to attain the feature representation that can be generally applied to other downstream tasks. Instead, our aim is to learn the feature representation that works best for a given retrieval dataset only. If this approach can be realised effectively, it can also be regarded as another kind of ``generally applicable'' solution but in the spirit of ``fits-itself-only.''

We observe that directly applying self-supervised learning (SSL) to learn feature representation from an image retrieval dataset does not work. Current SSL techniques work with object-level images, while the images in a retrieval dataset are often generic, containing multiple different objects within a same image. Blindly applying SSL in this case will not be able to learn any useful feature representation, as will be experimentally shown. Meanwhile, due to its unsupervised nature, a retrieval dataset usually lacks image- or object-level label information. In this case, it is hard to train any specific object detectors to obtain object regions. This situation drives us to utilise general-purpose unsupervised object proposal generators such as selective search~\cite{uijlings2013selective} or edge boxes~\cite{zitnick2014edge}. They do not need to be trained and only rely on low-level visual cues such as color, texture, and edge of an image to function. Surprisingly, it is observed that by merely using these generators to collect (noisy) object regions, we can enable SSL to successfully learn feature representation from each image retrieval benchmark dataset commonly used in the literature. Particularly, for the benchmark dataset {INSTRE}, which is not closely similar to ImageNet~\cite{deng2009imagenet} or the Google Landmarks~\cite{teichmann2019detect} dataset (usually used as an auxiliary dataset for building-based retrieval), the feature representation learned by the investigated approach outperforms all the existing methods. This is achieved without leveraging any pre-trained deep models or external auxiliary datasets, showing the potential of this approach.

On top of this, we investigate if the feature representation can be further boosted by mining image similarity information from the retrieval dataset, say, by diffusion process~\cite{zhou2004ranking}, query expansion, or region matching~\cite{razavian2016visual}. This investigation is motivated by two considerations. First, the literature on instance image retrieval has shown that these operations are effective to further improve retrieval. We are interested in examining if the SSL-learned representation also enjoys this property. More importantly, we believe that this investigation could enforce a tighter integration of the feature representation learned by SSL with a given retrieval dataset. After all, the primary function of SSL is to learn the intrinsic and generic invariances of images. We speculate that the SSL-learned representation has not well absorbed the underlying distribution information of the given dataset. Our investigation is positive. By mining image similarity to boost the learned feature representation, our method can compete with relevant state-of-the-art ones on most image retrieval benchmark datasets. 

Contributions of this work are summarised as follows. 
\begin{enumerate}
\item[1)] It explores a radical alternative to learn feature representation for instance image retrieval purely from within a retrieval dataset. Merely with existing self-supervised learning techniques, this work has been able to demonstrate the promising retrieval performance obtained by this approach. It has clear advantages of not requiring any pre-trained deep models or auxiliary datasets and being free of domain gap, while not pursuing feature representation that is generally applicable. 

\item[2)] To the best of our knowledge, this work is the first one conducting a thorough investigation on this approach to reveal its advantages and limitations. This approach could be a valuable option in terms of retrieval performance for many practical applications in which the datasets for retrieval have a static or stable data distribution. 

\item[3)] It shows that in the context of instance image retrieval, the feature representation learned by SSL can be further strengthened by absorbing additional intrinsic information of a dataset. This helps to achieve even more competitive performance with respect to relevant state-of-the-art methods, while fully maintaining the advantage of no need to access pre-trained deep models or auxiliary data.

\item[4)] Experimental study is conducted to demonstrate the interesting properties of this novel ``self-supervised learning + self-boosting'' framework for instance image retrieval. In addition, the expected limitation on the generalisation capability of the learned feature representation to other datasets is also carefully discussed.
\end{enumerate}

\section{Related Work}
\subsection{Instance image retrieval}
The state-of-the-art instance image retrieval is usually based on deep feature representations that are extracted with the deep models either pre-trained on image benchmark datasets or fine-tuned on external labelled data.  

This line of research begins with directly applying the CNN models pre-trained on ImageNet to perform retrieval~\cite{kalantidis2016cross,tolias2015particular}. Soon after, further fine-tuning a pre-trained deep model with an external labelled dataset having a similar nature as the given retrieval dataset becomes popular and attains even better retrieval performance. The work proposed by Babenko~\etal~\cite{babenko2014neural} is among the first ones of this category. They manually collect and annotate an external landmark dataset and use it to fine-tune the pre-trained deep model. The performance of retrieval is dramatically improved, especially on landmark retrieval tasks. To avoid costly manual annotation, Radenovi{\'c}~\etal~\cite{radenovic2018fine} utilise Structure-from-Motion (SfM) to generate more reliable image matching to guide the selection of training data from the external dataset~\cite{schonberger2015single} for fine-tuning. Also, Gordo~\etal~\cite{gordo2017end} use handcrafted local descriptors and spatial verification technique to remove outliers in the external dataset~\cite{babenko2014neural}, giving rise to a cleaned dataset for fine-tuning. Besides, some works in this category focus on providing a larger external dataset or more precise annotations~\cite{teichmann2019detect,noh2017large} to conduct fine-tuning, and others aim to develop better model architectures~\cite{cao2020unifying} or more effective loss functions~\cite{revaud2019learning,DBLP:conf/eccv/NgBTM20} for fine-tuning. 

As seen, accessing an external labelled dataset is a key factor to the success of these methods. This helps to align the deep model pre-trained on a generic benchmark dataset with a given retrieval dataset (say, landmark retrieval focused by the aforementioned works), leading to a more appropriate feature representation. Nevertheless, collecting an external relevant dataset and making it annotated, whether manually or algorithmically, could be awkward or expensive. Also, once a new retrieval task of different nature is encountered, the current external dataset may not be suitable anymore and a new dataset has to be collected and annotated again. These issues make this approach difficult to be widely implemented in practice. In this sense, an alternative free of the requirement to access pre-trained deep models or external labelled data becomes well-motivated and worth investigating.     

\begin{figure}[t]
\begin{center}
\includegraphics[width=1.0\linewidth]{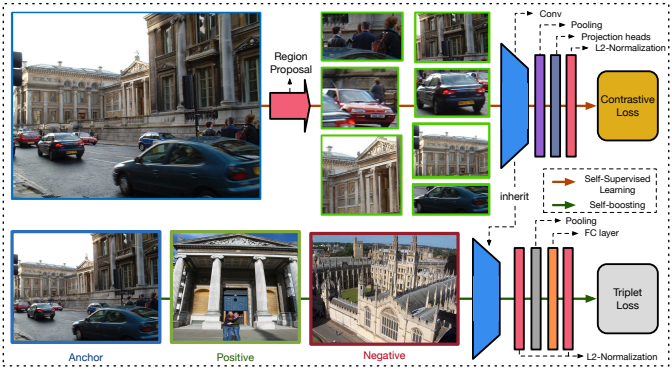}
\end{center}
   \caption{The two key parts of the proposed framework: 1) Learning feature representation purely from within a retrieval dataset by region-based self-supervised learning; and 2) Self-boosting the learned feature representation by exploiting intrinsic information of this retrieval dataset.}
\label{fig:pipeline}
\end{figure}

\subsection{Self-supervised learning (SSL)}

SSL has recently attracted intensive attention and exhibited promising results on learning feature representation~\cite{chen2020simple,he2020momentum,li2020prototypical,DBLP:conf/icml/ZbontarJMLD21,DBLP:conf/cvpr/ChenH21, DBLP:conf/nips/GrillSATRBDPGAP20}. Usually, a variety of different pretext tasks~\cite{DBLP:conf/iclr/GidarisSK18,DBLP:conf/cvpr/PathakGDDH17,DBLP:conf/eccv/ZhangIE16} are conducted on a large-scale generic unlabelled dataset to pre-train the model, and a down-stream task fine-tunes the model on a target dataset for specific applications. Among them, contrastive learning based methods show great potential and achieve significant performance gain. For example, Chen~\etal~\cite{chen2020simple} propose a simple framework consisting of data augmentation and contrastive learning (i.e., InfoNCE~\cite{oord2018representation}) to learn features based on instance invariance, demonstrating substantial improvement over previous methods. After that, He~\etal~\cite{he2020momentum} design a momentum framework to address the large memory cost and parameter updating issues. Meanwhile, other methods~\cite{DBLP:conf/cvpr/Chen000DLMX0021, DBLP:conf/eccv/TianKI20, DBLP:conf/icml/Henaff20, DBLP:conf/icml/ZbontarJMLD21} have been proposed to further improve the SSL performance. The effectiveness of SSL has been well verified on various tasks including image classification, detection, segmentation, and so on~\cite{jing2020self, jaiswal2021survey}.

Meanwhile, the potential of SSL to image retrieval has not been sufficiently investigated. In this work, we will utilise SSL to realise our investigation on learning purely from within a dataset. Note that our work conducts SSL in a special setting, that is, we do not assume the access to a large generic dataset in a pretext task but directly carry out SSL on the target dataset (i.e., the dataset of a given retrieval task). More importantly, as emphasised in the Introduction section, our goal of doing SSL is to learn a feature representation that can best work for a specific retrieval task, rather than learning a feature representation that is generally applicable to other down-stream tasks as focused by common SSL methods.   

\section{Proposed Method}

To investigate the question raised in the Introduction section, we propose a novel framework to learn feature representation for retrieval. It consists of two parts: 1) conducting self-supervised learning on a retrieval dataset to learn an initial feature representation; 2) boosting this representation by mining image similarity from this dataset. The two parts are presented as follows and they are also illustrated in Fig.~\ref{fig:pipeline}.

\subsection{Self-supervised learning on image regions}\label{sec:self}
The success of recent contrastive learning based SSL techniques largely relies on data augmentation~\cite{chen2020simple}. Commonly, a variety of transforms (\eg, random crop, color distortion, and Gaussian blur, etc.) are applied to an object-level image, and the loss enforces the learned feature representation of two transformed images to be close to each other. Clearly, this approach implicitly assumes that under these data augmentation operations, a transformed image will be able to retain the same instance. This assumption can be well met by object-level images. 

\begin{figure}[ht]
\centering
\footnotesize{
\begin{tabular}{c c c c}
\includegraphics[width=0.7in]{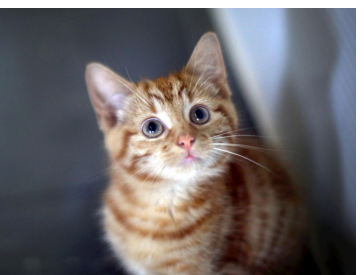}
& \includegraphics[width=0.7in]{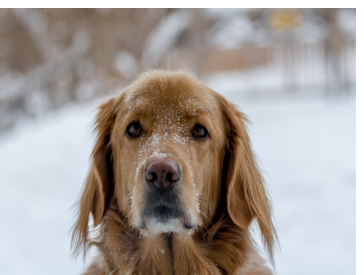} & \includegraphics[width=0.7in]{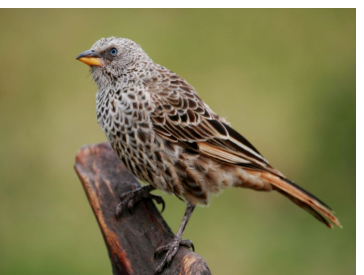} & \includegraphics[width=0.7in]{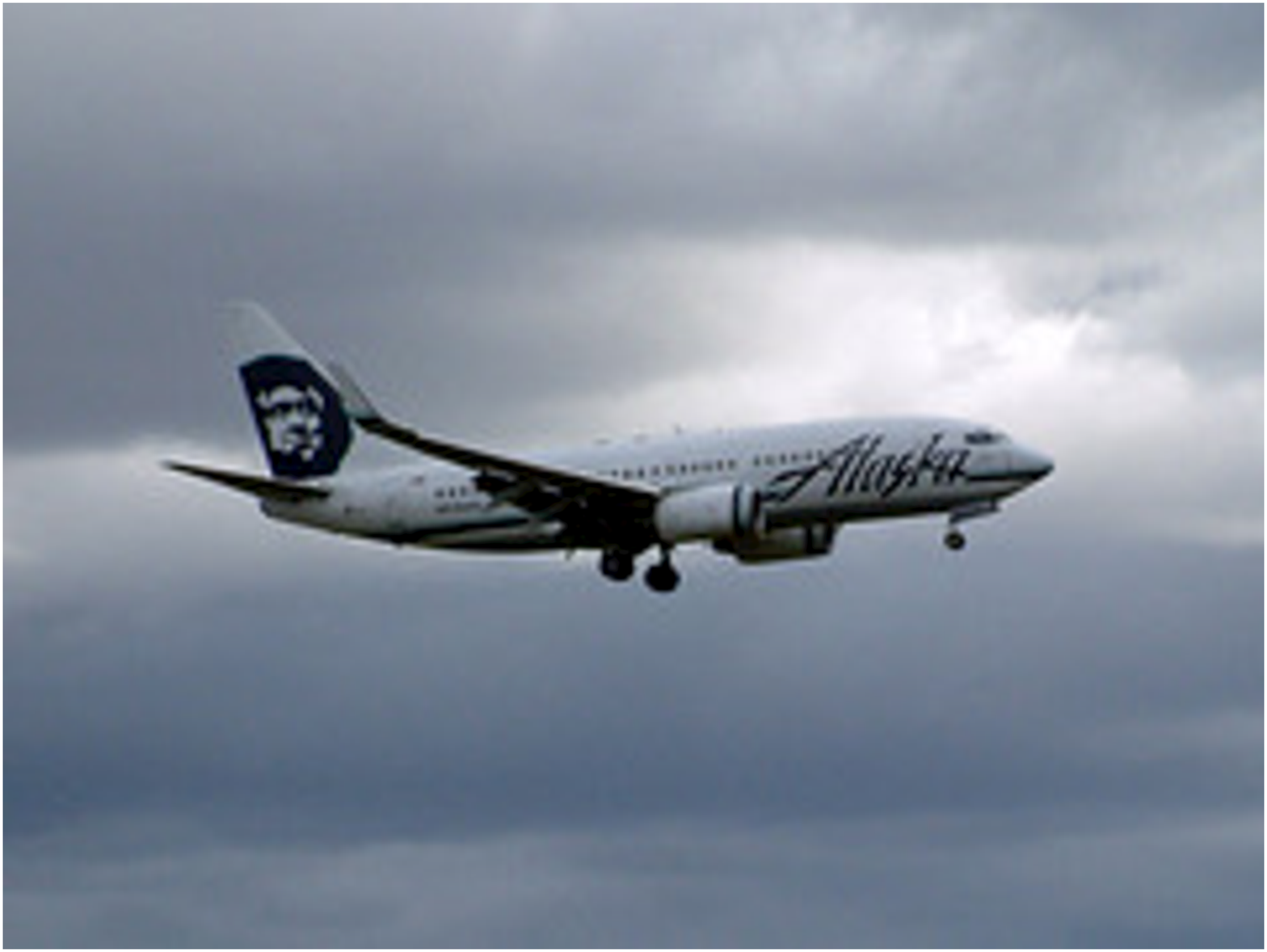} \\
\includegraphics[width=0.7in]{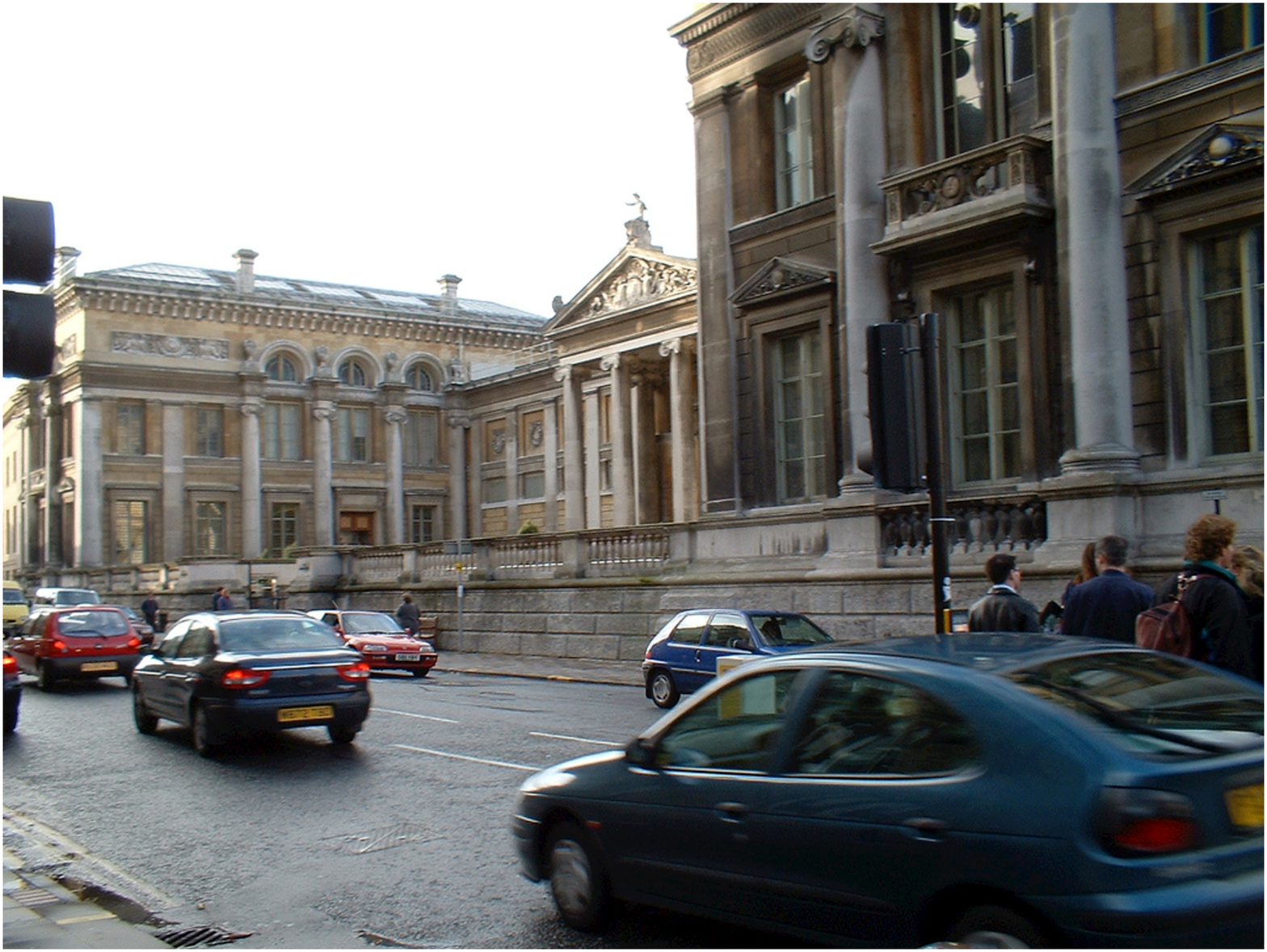} & \includegraphics[width=0.7in]{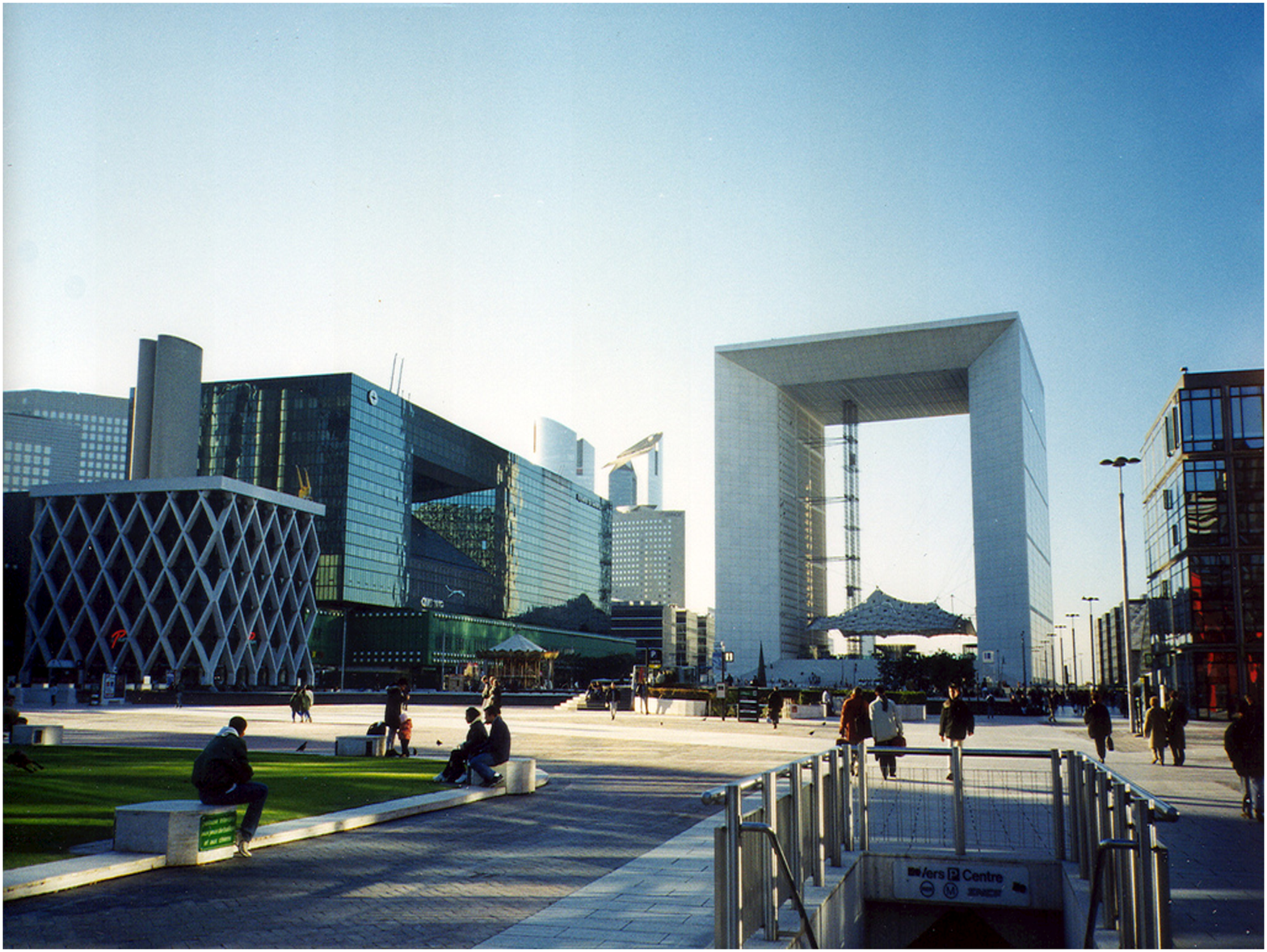} &\includegraphics[width=0.7in]{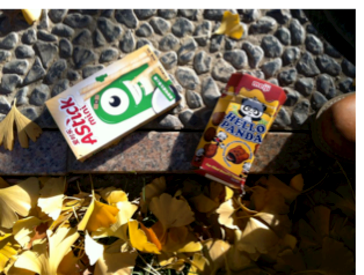} & \includegraphics[width=0.7in]{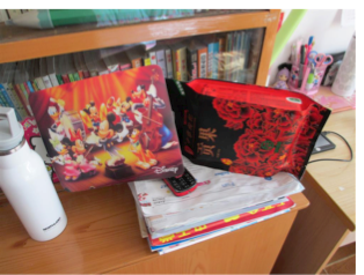}\\\
\end{tabular}
}
\caption{Example images in ImageNet~\cite{deng2009imagenet} (top) and the retrieval datasets (bottom) of $\mathcal R$Oxford, $\mathcal R$Paris~\cite{radenovic2018revisiting}, and INSTRE~\cite{wang2015instre}.}
\label{fig:demo}
\end{figure}

Nevertheless, datasets in the task of instance image retrieval are much less constrained. As a result, a single image could contain multiple instances of different object classes with complex background, as illustrated in Fig.~\ref{fig:demo}. Directly applying data augmentation to such an image can easily result in two transformed images that do not share the same object, which contradicts the implicit assumption in SSL. As will be experimentally demonstrated, existing SSL technique cannot effectively learn useful feature representation in this situation. 

Furthermore, it is not viable for us to train object detectors to obtain object regions from these images. This is because i) a dataset for retrieval usually does not have any label information and ii) the object classes presented in a retrieval dataset are often diverse and they cannot be precisely known without sufficiently inspecting all the images. In addition, a retrieval dataset may not be as large as the generic datasets conventionally used for SSL. This could limit the number of training samples available for SSL to effectively learn feature representation.\footnote{For example, the benchmark image retrieval datasets $\mathcal R$Oxford and $\mathcal R$Paris only contain $4,993$ and $6,322$ images. In contrast, the generic image datasets commonly used in the SSL tasks such as ImageNet1M and Instagram-1B contain about $1.28$ million and $1$ billion images.} 

To address this situation, we resort to general-purpose unsupervised object proposal generators (\eg, selective search~\cite{uijlings2013selective} and edge boxes~\cite{zitnick2014edge}) to discover object regions. Given an image, they output a set of bounding boxes indicating the regions that have a high possibility of containing an object. Note that these generators are unsupervised and work based on grouping superpixels or counting edge contours in an image. They are not deep models and do not need to be trained. In the literature, they have been widely used in various computer vision tasks~\cite{girshick2014rich,lin2017focal}.

With these object proposal generators, a sufficient number of regions will be generated for each image, providing enough training samples to train the SSL model, even when the size of the retrieval dataset is not large enough. By doing so, we change the training data from image-level to region-level to train SSL and the chance of resulting in two transformed data that contain different objects can be well reduced. Note that, many of the generated image regions do not really contain an object due to the limited performance of the general-purpose object proposal generators. These regions cannot be easily removed unless resorting to more sophisticated image recognition models, which is not in our focus. Nevertheless, our experimental study will show that we can still learn meaningful feature representation for retrieval by directly conducting SSL with the generated (noisy) regions. 

Then the application of SSL is straightforward. Formally, following the contrastive loss based SSL, we employ the InfoNCE loss~\cite{oord2018representation} to train a deep model to learn feature representation. Let ${\mathbf x}_i$ and ${\mathbf x}'_i$ denote the two transformed variants from the $i$th image region generated from the images in a retrieval dataset. The target is to learn an embedding function ${\mathbf z} = f({\mathbf x})$ to make sure ${\mathbf z}_i$ and ${\mathbf z}'_i$ to be close to each other. The loss function is written as
\begin{equation}\label{eq:info}
   \ell_{\operatorname{Info}}({\mathbf z}_i,{\mathbf z}'_i) = -\log\frac{\exp{(\operatorname{sim}({\mathbf z}_i,{\mathbf z}'_i) \slash \tau)}}{\sum_{k=1}^{N} \mathbbm{1}_{[k \neq i]} \exp{(\operatorname{sim}({\mathbf z}_i, {\mathbf z}'_k) \slash \tau)}},
\end{equation}
where $\operatorname{sim}(\cdot,\cdot)$ is a similarity measure, $\mathbbm{1}_{[k \neq i]} \in \{0,1\}$ is an indicator function, $N$ is the batch size, and $\tau$ is a temperature hyper-parameter. To investigate the potential of existing SSL techniques to our task, we just follow the common protocols to train this deep model without further modification. After training, we simply use the obtained deep model to extract a global feature representation directly from each whole image to conduct retrieval.

\subsection{Self-boosting of the learned representation}\label{sec:self-boosting}
As will be shown, the feature representation learned via SSL in Section~\ref{sec:self} (will be called ``initial representation'' from now on) has been able to achieve quality retrieval on a given dataset. Nevertheless, there is still room to improve by considering the following two issues. First, the goal of SSL is to learn a feature representation that can be generally applied to other down-stream tasks. In contrast, our goal is to make it capable in correctly ranking images of the given dataset for a query. In other words, the feature representation learned via SSL does not directly serve the purpose of retrieval on the given dataset. For example, it is well realised that the current SSL techniques suffer the false positive issue~\cite{chuang2020debiased}. This could affect its performance on image retrieval. Second, the intrinsic information of the given dataset has not been fully exploited, while this is proven to be effective to further improve retrieval in the literature. Motivated by these, we fine-tune the initial representation learned by SSL with image similarity mined from the given retrieval dataset. Because this mining process itself is based on the learned initial feature representation, we call this part ``self-boosting of the learned representation.''

Various ways to mine image similarity have been seen in the literature of image retrieval. For example, Radenovi{\'c}~\etal~\cite{radenovic2018fine} utilise {Bag-of-Words} descriptors and Structure from Motion to infer image similarity for training data selection. Gordo~\etal~\cite{gordo2017end} use {handcrafted local descriptors} to do this for outlier cleaning. In addition, Iscen~\etal~\cite{iscen2018mining} and Zhao~\etal~\cite{zhao2018modelling} utilise diffusion process to obtain image similarity by considering the underlying feature distribution of the images in a dataset. All the mined image similarity can be turned into a kind of ``pseudo-labels'' to fine-tune a learned model. In principle, our work can utilise any one of the above methods to generate the pseudo-labels for self-boosting.

In sum, we mine image similarity entirely based on the initial feature representation learned by SSL. Pseudo-labels are then generated to fine-tune the initial representation model. Specifically, we formulate self-boosting as a commonly seen metric learning task, in which the pseudo-labels take the form of sample triplets (i.e., anchor, positive, and negative). This is to make the initial feature representation assimilate the mined image similarity information to become more capable for retrieval on the given dataset. 

\subsubsection{Pseudo-label generation} We generate training triplets by following the literature~\cite{iscen2018mining}. Let's assume that a new similarity for each pair of images in the given dataset for retrieval has been obtained by a certain mining method mentioned above, for example, diffusion process. According to the new similarity, for each anchor image ${\mathbf x}_a$, the ranking list consisting of its top-$k$ nearest neighbours is denoted by ${\mathbb R}_k({\mathbf x}_a)$. Also, the ranking list obtained based on the initial feature representation and Euclidean distance is denoted by ${\mathbb R}_k^e({\mathbf x}_a)$. The two ranking lists are compared and the discrepancy is used to choose the positives and negatives. Formally, with a given anchor ${\mathbf x}_a$, its positives and negatives are defined as\footnote{This training data selection approach assumes that with a top-$k$ ranking list for an image ${\mathbf x}_a$, the new (more reliable) similarity measure could retrieve more hard positives $P^{+}({\mathbf x}_a)$ which cannot be retrieved by Euclidean distance and that the initial ranking list obtained via Euclidean distance usually retrieves more false positives $P^{-}({\mathbf x}_a)$. Detailed explanation can be found in~\cite{iscen2018mining}.}
\begin{equation}\label{eq:pn}
\left\{
    \begin{array}{lr}
    P^{+}({\mathbf x}_a) = \{{\mathbf x} \in {\mathbb R}_k({\mathbf x}_a) \setminus {\mathbb R}_k^e({\mathbf x}_a)\}. \\
    P^{-}({\mathbf x}_a) = \{{\mathbf x} \in {\mathbb R}_k^e({\mathbf x}_a) \setminus {\mathbb R}_k({\mathbf x}_a)\}.
    \end{array}
\right.
\end{equation}
As for anchor selection, we select the images corresponding to the highest local density in the initial feature representation space as the anchors. This can be readily implemented by conducting the mean shift clustering~\cite{cheng1995mean} on the retrieval dataset, and the images that are nearest to the cluster centers are chosen as the anchors.

\subsubsection{Boosting the initial representation} Now we are ready to further boost the initial feature representation model learned via SSL for the given retrieval task. To help the initial model to assimilate the information on the mined image similarity, one fully connected layer is appended to the pooling layer while all the convolutional layers are frozen. This setting is commonly used in the literature when
a self-supervised pre-trained model is fine-tuned~\cite{chen2020simple, he2020momentum}. {By doing so, the original similarity of the images could be largely maintained and the new information on image similarity could be effectively absorbed by the added fully connected layer.} The output of this fully connected layer is regarded as the boosted feature representation for an image. Let's denote the whole embedding process from an image $\mathbf x$ to the boosted feature representation $\mathbf z$ as ${\mathbf z} = f_b({\mathbf x})$. With the triplets, a variety of losses, such as contrastive loss~\cite{radenovic2018fine}, listwise average prevision loss~\cite{revaud2019learning}, Quadruple loss~\cite{DBLP:conf/cvpr/ChenCZH17}, etc. could be utilised. Here, we simply adopt a commonly used triplet loss as follows to train the model 
\begin{equation}\label{eq:triplet}
\begin{aligned}
    \ell({\mathbf x}_a, {\mathbf x}^+, {\mathbf x}^-) = \max(m + {\|f_b({\mathbf x}_a) - f_b({\mathbf x}^+)\|}^2 \\
    - {\|f_b({\mathbf x}_a) - f_b({\mathbf x}^-)\|}^2, 0),
\end{aligned}
\end{equation}
where $m$ is the margin and ${\mathbf x}_a$, ${\mathbf x}^+$ (${\mathbf x}^+ \in P^{+}({\mathbf x}_a)$), and ${\mathbf x}^-$ (${\mathbf x}^- \in P^{-}({\mathbf x}_a)$) are the anchor, the positive, and the negative, respectively. Note that all ${\mathbf x}_a$, ${\mathbf x}^+$, and ${\mathbf x}^-$  are whole images instead of the image regions used in the SSL process in Section~\ref{sec:self}. 

\section{Experimental Result}
This section first introduces the experimental settings and the implementation of the proposed framework. After that, it demonstrates the retrieval performance obtained by the initial feature representation learned via SSL and then the performance attained after self-boosting. At last, ablation study is conducted.

\subsection{Experimental Settings}

\subsubsection{Dataset and evaluation} The proposed framework is tested on five benchmark datasets commonly used for instance image retrieval, \ie, INSTRE~\cite{wang2015instre}, $\mathcal R$Oxford, $\mathcal R$Paris, $\mathcal R$Oxford+$\mathcal R$1M distractors, and $\mathcal R$Paris+$\mathcal R$1M distractors~\cite{radenovic2018revisiting}. INSTRE contains 28,543 images depicting various objects such as toys, books, and logos with natural backgrounds. Following the protocol in~\cite{iscen2017efficient}, we partition this dataset into 27,293 images for retrieval and 1,250 images as queries. The rest four datasets are all landmark based. $\mathcal R$Oxford and $\mathcal R$Paris have 4,993 and 6,322 images of buildings and landmarks collected from Oxford and Paris, respectively, with 70 additional query images each. They are the improved versions of the Oxford5K~\cite{DBLP:conf/cvpr/PhilbinCISZ07} and Paris6K~\cite{DBLP:conf/cvpr/PhilbinCISZ08} datasets widely used in previous retrieval methods. Each image in $\mathcal R$Oxford and $\mathcal R$Paris is labelled as ``easy'', ``hard'', ``unclear'', or ``negative'' upon its quality of depicting a query instance~\cite{radenovic2018revisiting}. $\mathcal R$1M contains one million distractor images used to test the robustness of a retrieval method for large datasets. Following the literature, two evaluation settings of ``Medium'' and ``Hard'' are used for the landmark-based datasets. For each setting, the mean average precision (mAP) evaluates the retrieval for the predefined query images. We highlight that no query images are used in any place of this work to learn feature representation. 

\subsubsection{Implementation}
Following the state-of-the-art retrieval methods, our framework is built upon the ResNet101 model~\cite{DBLP:conf/cvpr/HeZRS16}.

\noindent\textit{SSL training part:} Given a benchmark retrieval dataset, a general-purpose unsupervised object proposal generator is used to search for object regions from each image. The regions with $95$\% overlapping areas are merged and the regions having either side shorter than 100 pixels are discarded. We build our self-supervised learning upon the framework of MoCo~\cite{he2020momentum} and set all the parameters as in the work of MoCo-v2.\footnote{Due to the limitation of computational resource, other SSL methods such as SimCLR~\cite{chen2020simple}, BYOL~\cite{DBLP:conf/nips/GrillSATRBDPGAP20}, and PCL~\cite{li2020prototypical} currently are not investigated for our framework. Based on their competitive SSL performance, we believe that they could lead to similar or better retrieval than MoCo~\cite{he2020momentum}.} After the convolutional layers, a global average pooling and two projection head layers are deployed to embed an input image region into a $128$-dimensional feature space in which the InfoNCE loss in Eq.~(\ref{eq:info}) is evaluated. After the training loss converges, the obtained model is called the initial feature representation model in this experiment. 

\noindent\textit{Self-boosting part:} Diffusion process, is used to mine image similarity within the given benchmark retrieval dataset. Advanced variants of diffusion methods~\cite{DBLP:conf/aaai/YangHMLS19,DBLP:conf/cvpr/IscenATFC18} have been developed, including those computationally more efficient in dealing with large image datasets. Since our focus is on validating the proposed framework, we just use the basic diffusion process method~\cite{zhou2004ranking} in this experiment. Specifically, the diffusion process is conducted based on the initial feature representations extracted from whole images to obtain new image similarity. For the $k$ in Eq.(\ref{eq:pn}), its value is empirically chosen for an anchor ${\mathbf x}_a$ as the one that makes the two ranking lists, ${\mathbb R}_k({\mathbf x}_a)$ and ${\mathbb R}_k^e({\mathbf x}_a)$, maximally dissimilar. The margin $m$ in Eq.(\ref{eq:triplet}) is empirically set as $0.2$ always. To conduct training, we keep all the layers before the global pooling layer used in SSL, and add a fully connected layer after them, which is the only layer used for training to obtain the self-boosted model. 

\begin{table}[t]
\centering{
\caption{Comparison with the state-of-the-art methods in terms of retrieval performance (in mAP). ResNet101 is used as feature extractor backbone in the table, and the ones used in the compared methods are pre-trained on ImageNet~\cite{deng2009imagenet}. ``INS'', ``$\mathcal R$Oxf'', and ``$\mathcal R$Par'' represent INSTRE, $\mathcal R$Oxford, and $\mathcal R$Paris in short. $ \dagger$ denotes that the performance is obtained via running the source code released by the authors; $\ddagger$ means the results are re-implemented by~\cite{radenovic2018revisiting}. $\times$ means that the corresponding method does not use this dataset. ``F-tune'' lists the external datasets used for fine-tuning.}
\label{tab:sota}
\begin{tabular}{l c c c c c c}
 \hline
 \multirow{2}{*}{Method} & \multirow{2}{*}{F-tune} & \multirow{2}{*}{INS} & \multicolumn{2}{c}{Medium} & \multicolumn{2}{c}{Hard} \\
 \cline{4-7}
  & & & $\mathcal R$Oxf & $\mathcal R$Par & $\mathcal R$Oxf & $\mathcal R$Par \\
 \hline
 Crow~\cite{kalantidis2016cross}$\ddagger$ & $\times$  & 26.7$\dagger$ & 41.4 & 62.9 & 13.9 & 36.9 \\
 R-MAC~\cite{tolias2015particular}$\ddagger$ & $\times$  & 43.0$\dagger$ & 42.5 & 66.2 & 12.0 & 40.9 \\
 R-MAC~\cite{gordo2017end} & \cite{babenko2014neural} & - & 60.9 & 78.9 & 32.4 & 59.4 \\
 GeM~\cite{radenovic2018fine} & \cite{schonberger2015single} & 69.0$\dagger$ & 64.7 & 77.2 & 38.5 & 56.3\\
 GeM+AP~\cite{revaud2019learning} & \cite{babenko2014neural} & 33.5$\dagger$ & 67.5 & 80.1 & 42.8 & 60.5\\
 GeM~\cite{radenovic2018fine} & \cite{teichmann2019detect} & 58.8$\dagger$ & 67.3 & 80.6 & 44.3 & 61.5 \\
 SOLAR~\cite{DBLP:conf/eccv/NgBTM20} & \cite{teichmann2019detect} & - & 69.9 & 81.6 & 47.9 & 64.5 \\
 DELG~\cite{cao2020unifying}  & \cite{noh2017large} & - & 73.2 & 82.4 & \textbf{51.2} & 64.7 \\
 \hline
 ${SSL}_{eb}$(Ours) & $\times$ & 92.2 & 68.2 & 78.0 & 43.0 & 56.9 \\
 ${SSL}_{ss}$(Ours) & $\times$ & 92.0 & 69.1 & 81.5 & 44.8 & 63.0 \\
 ${SSL}_{eb}$+$B_d$(Ours) & $\times$ & \textbf{92.5} & \textbf{73.5} & \textbf{86.4} & 51.0 & 69.3 \\
 ${SSL}_{ss}$+$B_d$(Ours) & $\times$ & 92.3 & 71.5 & 84.8 & 46.1 & \textbf{70.7} \\
 \hline
\end{tabular}}
\end{table}

\noindent\textit{Retrieval part:} 
To extract the initial feature representation, we simply add the Crow pooling~\cite{kalantidis2016cross} layer after the convolutional layers trained by SSL only (i.e., before self-boosting is performed). To extract the boosted feature representation, we directly utilise the model fine-tuned by the self-boosting process to extract the features output by the fully connected layer. During this process, the same Crow pooling~\cite{kalantidis2016cross} is used. For either case, each whole image in the given benchmark dataset is used as an input. The obtained global feature vector is $l_2$-normalised to measure the image similarity with Euclidean distance. Following existing methods in the literature, the dataset images and the queries are all resized to have the longer side of 1,024 pixels, and only this single scale is used to extract features in our framework.

\subsection{Comparison with the state-of-the-art}\label{sec:comparison}

This section compares our framework with other relevant state-of-the-art retrieval methods that utilise a pre-trained deep model and/or a labelled external dataset. Specifically, the methods relying on fine-tuning a pre-trained deep model with an external semi-automatically or manually labeled dataset are compared, including R-MAC~\cite{gordo2017end}, GeM~\cite{radenovic2018fine}, GeM+AP~\cite{revaud2019learning}, SOLAR~\cite{DBLP:conf/eccv/NgBTM20} and DELG~\cite{cao2020unifying}. In addition, to better demonstrate that our framework does not need to rely on any external supervised deep learning models, the methods (\eg, Crow~\cite{kalantidis2016cross} and R-MAC~\cite{tolias2015particular}) that directly use an ImageNet pre-trained deep model are also included in the comparison. To make the comparison clear, we list the external datasets used by the compared methods for fine-tuning (in column ``F-tune'') in Table~\ref{tab:sota}. As for our framework, we report its retrieval performance obtained by the initial feature representation learned by SSL, with either selective search~\cite{uijlings2013selective} (reported as ${SSL}_{ss}$) or edge boxes~\cite{zitnick2014edge} (reported as ${SSL}_{eb}$) as the object proposal generator. For our retrieval performance after self-boosting, the corresponding two results are reported as ${SSL}_{ss}$ + $B_d$ and ${SSL}_{eb}$ + $B_d$, respectively.

Table~\ref{tab:sota} has two sections. The top section consists of the compared methods. It can be seen that compared with those directly using an ImageNet pre-trained model (i.e, the first two methods), the methods fine-tuned with external datasets (i.e., the rest six methods) indeed achieve much better retrieval. The bottom section consists of the proposed methods which do not need to access any pre-trained deep models or any external labeled datasets. Surprisingly, the initial feature representation simply trained by SSL on each dataset individually has already obtained promising retrieval on all benchmark datasets. As seen from the rows of ${SSL}_{eb}$ and ${SSL}_{ss}$, their mAP values have been comparable or even higher than some of the rest six methods in the top section.  

The merit of our framework can be best observed from the benchmark dataset INSTRE, on which our method outperforms those in comparison by a large margin (92.0+ vs. 69.0). This significant difference is explained as follows. The external datasets used by the compared methods are landmark based, which has a different nature from INSTRE. As a result, although these methods excel on $\mathcal R$Oxford and $\mathcal R$Paris (which are also landmark based) by leveraging the external dataset, they cannot effectively generalise this success to INSTRE due to the domain gap between the two datasets. This is exactly the shortcoming of this kind of approach we point out previously. In addition, this shortcoming can even be observed for the models pre-trained on the generic ImageNet dataset and do not involve fine-tuning with external datasets. They correspond to the first two methods in the table and their performance on INSTRE is clearly not good either. In contrast, our feature representation is learned purely from within the INSTRE dataset. It can well reflect the nature of this specific dataset and is therefore completely free of this domain gap issue. This is why it achieves much better retrieval on the INSTRE dataset.

The result of applying self-boosting is reported in the rows of ${SSL}_{eb}$ + $B_d$ and ${SSL}_{ss}$ + $B_d$ in Table~\ref{tab:sota}. As seen, on all the datasets the proposed framework attains even better retrieval than the case using the initial feature representation learned by SSL. This indicates that the intrinsic information of a given retrieval datasets is helpful for the SSL-learned representation, and it is effectively absorbed by our model via the self-boosting step
. These results show the great potential of the proposed SSL-based framework in directly learning feature representation for the retrieval on a given dataset. 

\begin{table}[ht]
{\small
\begin{center}
\caption{Comparison with the state-of-the-art methods in terms of retrieval performance (in mAP) on $\mathcal R$Oxford and $\mathcal R$Paris with $+\mathcal R$1M distractors. ``M'' and ``H'' denote Medium and Hard settings. All the rest abbreviations are used the same as in Table~\ref{tab:sota}.}
\label{tab:sota-1m}
\begin{tabular}{l c c c c c}
 \hline
 \multirow{2}{*}{Method} & \multirow{2}{*}{F-tune} & \multicolumn{2}{c}{$\mathcal R$Oxf$+$1M} & \multicolumn{2}{c}{$\mathcal R$Par$+$1M} \\
 \cline{3-6}
  & & M & H & M & H \\
 \hline
 Crow~\cite{kalantidis2016cross}$\ddagger$ & $\times$  & 22.5 & 3.0 & 34.1 & 10.3 \\
 R-MAC~\cite{tolias2015particular}$\ddagger$ & $\times$ & 21.7 & 1.7 & 39.9 & 14.8 \\
 R-MAC~\cite{gordo2017end} & \cite{babenko2014neural} & 39.3 & 12.5 & 54.8 & 28.0 \\
 GeM~\cite{radenovic2018fine} & \cite{schonberger2015single} & 45.2 & 19.9 & 52.3 & 24.7 \\
 GeM+AP~\cite{revaud2019learning} & \cite{babenko2014neural} & 47.5 & 23.2 & 52.5 & 25.1 \\
 GeM~\cite{radenovic2018fine} & \cite{teichmann2019detect} & 49.5 & 25.7 & 57.3 & 29.8 \\
 SOLAR~\cite{DBLP:conf/eccv/NgBTM20} & \cite{teichmann2019detect} & 53.5 & 29.9 & 59.2 & 33.4 \\
 DELG~\cite{cao2020unifying} & \cite{noh2017large} & 54.8 & 30.3 & 61.8 & 35.5 \\
 \hline
 \multicolumn{6}{l}{Setting I: Learning on $\mathcal R$Oxf (or $\mathcal R$Par) and testing on $\mathcal R$Oxf$+$1M} \\
 \multicolumn{6}{l}{(or $\mathcal R$Par$+$1M).} \\
 \hline
 ${SSL}_{eb}$(Ours) & $\times$ & 47.1 & 20.9 & 48.1 & 18.1 \\
 ${SSL}_{ss}$(Ours) & $\times$ & 50.3 & 27.7 & 53.1 & 27.9 \\
 ${SSL}_{eb}$+$B_d$(Ours) & $\times$ & 59.8 & 36.2 & 72.2 & 48.9 \\
 ${SSL}_{ss}$+$B_d$(Ours) & $\times$ & \textbf{60.8} & \textbf{36.4} & 70.6 & 49.9\\
 \hline
 \multicolumn{6}{l}{Setting II: Learning on $\mathcal R$Oxf$+$1M (or $\mathcal R$Par$+$1M) and testing on} \\
 \multicolumn{6}{l}{the same dataset.} \\
 \hline
 ${SSL}_{eb}$(Ours) & $\times$ & 32.4 & 12.8 & 72.7 & 53.8 \\
 ${SSL}_{ss}$(Ours) & $\times$ & 32.7 & 12.7 & 74.4 & 60.2 \\
 ${SSL}_{eb}$+$B_d$(Ours) & $\times$ & 32.7 & 14.1 & 73.1 & 53.9 \\
 ${SSL}_{ss}$+$B_d$(Ours) & $\times$ & 31.1 & 12.8 & \textbf{74.5} & \textbf{61.9} \\
 \hline
\end{tabular}
\end{center}
}
\end{table}

\subsection{Ablation study}

\subsubsection{Robustness to noisy images}
Since the feature representation in this work is learned purely from a given dataset, we are interested in understanding its robustness with respect to the presence of noisy images. Following previous methods~\cite{cao2020unifying, DBLP:conf/eccv/NgBTM20, revaud2019learning, radenovic2018fine}, we use $\mathcal R$1M dataset~\cite{radenovic2018revisiting}, which consists of one million distractor images, as the noisy images. 

To be comprehensive, this experimental study considers two different settings: 1) The $\mathcal R$1M dataset is only merged into $\mathcal R$Oxford (or $\mathcal R$Paris) at the retrieval stage. That is, it is not seen when the feature representation is learned; 2) The $\mathcal R$1M dataset is merged into $\mathcal R$Oxford (or $\mathcal R$Paris) from the beginning to make both learning and retrieval conducted on $\mathcal R$Oxford+$\mathcal R$1M (or $\mathcal R$Paris+$\mathcal R$1M). The former investigates the robustness of the learned representation against unseen noisy images, while the latter investigates the robustness of learning feature representation in the presence of noisy images.  

Table~\ref{tab:sota-1m} reports the retrieval results. For comparison, all the methods compared in the above section are also applied and their results are provided in the first part of this table. The results for the two settings considered in this experiment are shown under ``Setting I'' and ``Setting II''. Interestingly, the retrieval performance shows a dramatic difference between the two settings. 

For the first setting, when self-boosting is not applied, our method shows similar performance drop (comparing with the performance when no distractors are involved in Table~\ref{tab:sota}) as the other methods in comparison. However, after conducting self-boosting, its retrieval performance is dramatically increased and outperforms all the others by a large margin (\eg, 72.2 vs. 61.8 on $\mathcal R$Paris$+\mathcal R$1M). This result suggests that the self-boosting step helps capture more intrinsic information from the retrieval dataset. And it learns better feature representations that could somehow have the effect of ``distinguishing'' the distractors from the original images in $\mathcal R$Oxford or $\mathcal R$Paris.



As seen for the second setting, the retrieval performance of the learned representation is adversely impacted by the presence of distractors. Nevertheless, the extent of this impact is different for the two datasets. The retrieval performance drops greatly on $\mathcal R$Oxford+$\mathcal R$1M (i.e., $20\sim{30}$ percentage points lower than Setting I). Differently, the performance increases slightly on $\mathcal R$Paris+$\mathcal R$1M (i.e., $2\sim{10}$ percentage points higher than Setting I). Actually, it outperforms all the compared methods listed in the first part by a large margin on this dataset (e.g., $61.9$ vs. $35.5$ as shown in the last column of Table~\ref{tab:sota-1m}). We speculate that the different change of retrieval performance on the two datasets could be related to their domain gap from $\mathcal R$1M. Note that the dataset of $\mathcal R$1M is much larger and more diverse than $\mathcal R$Oxford and $\mathcal R$Paris. This means the feature representation learned from within $\mathcal R$Oxford+$\mathcal R$1M or $\mathcal R$Paris+$\mathcal R$1M is overall determined by $\mathcal R$1M. As analysed in the Appendix, we found that the domain gap between $\mathcal R$Paris and $\mathcal R$1M is smaller, which makes the learned feature representations more effective to maintain the retrieval performance.

\begin{table}
\begin{center}
\caption{Retrieval performance (by mAP) of the feature representations learned from the training set under different sampling ratios on $\mathcal R$Paris.}
\label{tab:overfitting}
\begin{tabular}{l c c c c c}
\hline
Set\textbackslash Sampling Ratio & 50\% & 60\% & 70\% & 80\% & 90\% \\
 \hline
 \multicolumn{6}{c}{Medium Setting} \\
 \hline
 Training set & 75.4 & 76.2 & 76.4 & 76.7 & 77.2\\
 Test set & 74.4 & 76.0 & 78.4 & 78.4 & 79.6 \\
 \hline
 \multicolumn{6}{c}{Hard Setting} \\
 \hline
 Training set & 55.0 & 55.1 & 56.1 & 54.9 & 56.7\\
 Test set & 51.4 & 52.8 & 55.3 & 55.7 & 60.9 \\
 \hline
\end{tabular}
\end{center}
\end{table}

\subsubsection{Generalisation}\label{sec:overfitting}
When training and test are conducted on the same dataset, an interesting issue to investigate is the generalisation of the feature representations learned by our framework. In this part, we analyse it in two aspects including: 1) the generalisation to a set of new images with a similar data distribution as the training dataset; 2) the generalisation to a set of new images with a data distribution different from that of the training dataset. To be accurate and fair, our feature representation learned only by SSL (with edge boxes as the region proposal generator) is used, and the self-boosting step is not applied for the following study.

For the first aspect, we sample part of the dataset as the training set to learn the feature representations and conduct retrieval on the rest part of the dataset as the test set. In particular, $\mathcal R$Paris is selected to implement this experiment since it has a relatively balanced object class distribution. The sampling is uniformly conducted for each building class in the dataset to avoid altering the data distribution across the training and test sets.\footnote{As to the query-irrelevant images, we regard them as a specific class, and conduct sampling in the same way as other query-relevant classes} The performance is demonstrated in Table~\ref{tab:overfitting}. As seen, for both settings, no matter which sampling ratio is used, the retrieval performance on the training set and the test set are always similar. It suggests that the feature representations learned by our method could be well generalised to the new unseen data when it has a similar data distribution as the seen (training) data. In addition, we repeat the experiment by conducting sampling at the ratio of 50\% by five times individually to reduce the randomness of the performance in Table~\ref{tab:overfitting}. The obtained performance are $75.5 \pm 1.5$ (M) / $53.7 \pm 3.1$ (H) on the training set and $75.1 \pm 1.5$ (M) / $53.4 \pm 3.5$ (H) on the test set. As seen, the retrieval performance on both Medium and Hard settings is still comparable across the training and test sets, which further validates the aforementioned conclusion on generalisation.

\begin{table}
\begin{center}
\caption{Investigation of the generalisation capability of the feature representation learned by SSL in the proposed framework. The representation learned on one dataset is tested on another. The same model architecture pre-trained on ImageNet and Google Landmark-v2-clean~\cite{DBLP:conf/cvpr/WeyandACS20} (referred as GL-v2-clean here) is also included.}
\label{tab:generality}
\begin{tabular}{l| c c c}
\hline
 Learned on \textbackslash Tested on & INSTRE & $\mathcal R$Oxford & $\mathcal R$Paris \\
 \hline
 INSTRE &  \textbf{92.2} &  24.2 & 55.5 \\
 $\mathcal R$Oxford & 12.0 & \textbf{68.2} & 45.8\\
 $\mathcal R$Paris & 13.2 & 22.3 & \textbf{78.0} \\
 \hline
 ImageNet (Supervised) & 26.7 & 18.9 & 39.6 \\
 GL-v2-clean (SSL only)~\cite{DBLP:conf/cvpr/WeyandACS20} & 10.4 & 22.7 & 46.0 \\
 \hline
\end{tabular}
\end{center}
\end{table}

For the second aspect, we apply the feature representation learned by our framework from one retrieval dataset to another dataset to conduct retrieval. As seen in Table~\ref{tab:generality}, when the same dataset is used for learning and retrieval, the mAP values are clearly high. For instance, on INSTRE our retrieval performance could reach 92.2. However, when applying the same feature representation to the datasets $\mathcal R$Oxford and $\mathcal R$Paris of a different (landmark-based) nature, the obtained mAP values (24.2 and 55.5) are much lower than what we achieve by using the feature representation directly learned from the same dataset (68.2 and 78.0). When the feature representation learned on $\mathcal R$Oxford or $\mathcal R$Paris is applied to other datasets, a similar result can be observed. In addition, to provide a reference, we investigate the generalisation ability of the feature representation obtained by a ResNet101 model trained on ImageNet in a supervised manner (released by the PyTorch official website~\cite{paszke2019pytorch}). The result is listed at the second last row of Table~\ref{tab:generality}. As seen, it does not generalise well either. This observation is not new to the image retrieval community, and this is why various ways from PCA whitening to external dataset based fine-tuning are used in the current retrieval methods to adapt the ImageNet pre-trained models. In this sense, the proposed framework is a kind of radical alternative that learns feature representation directly from the retrieval dataset itself, instead of adapting a pre-trained representation to this dataset. Furthermore, we apply SSL on Google Landmark-v2 Clean~\cite{DBLP:conf/cvpr/WeyandACS20}, a commonly used external dataset for fine-tuning pre-trained models for image retrieval, to learn feature representations and conduct retrieval on the above three datasets. As shown in the last row of Table~\ref{tab:generality}, without the help of its class annotations, this landmark-relevant external dataset can only provide limited benefit to retrieval on $\mathcal R$Oxford and $\mathcal R$Paris. This demonstrates that, even though the external dataset shares a similar nature with the target retrieval dataset, the generalisation of the learned feature representations can still not be guaranteed. 

To sum up, the feature representation learned by our framework could be well generalised to a dataset with a similar data distribution as the training data. Meanwhile, it does not necessarily generalise to a dataset of a nature different from the training data. However, we emphasise that this representation works surprisingly well on the retrieval dataset from which it is learned. This is consistent with our goal.

\subsubsection{The effect of object proposal generator}
The proposed framework utilises general-purpose unsupervised object proposal generators to discover object regions for SSL-based feature representation learning. To show its effect, this experiment tests three different object proposal generators for this purpose, with an additional case in which whole images are directly used. The three generators are selective search~\cite{uijlings2013selective}, edge boxes~\cite{zitnick2014edge}, and a grid-partition scheme~\cite{tolias2015particular}. The last one uniformly samples image regions of various sizes based on a grid predefined for an image and has been used in R-MAC~\cite{tolias2015particular} to obtain object regions. The only parameter in this scheme is the scale $l$ and we set $l=6$ in this experiment to make sure the number of the generated regions for each image is comparable to the others. After obtaining the object region proposals, the overlapping cases and the smaller-sized proposals will be processed in the way as aforementioned in the main text. 

\begin{table}[ht]
\begin{center}
\caption{Retrieval performance (in mAP) with feature representations learned by SSL based on the object regions obtained by different object proposal generators. Crow pooling~\cite{kalantidis2016cross} is uniformly used and the SSL is applied to extract the feature representation. ``WI'', ``GP'', ``EB'', and ``SS'' denote whole image, grid-partition, edge boxes, and selective search respectively. Epochs and \# Regions represent the number of training epochs for convergence and the number of the obtained object regions per image.}
\label{tab:object-discovery}
\begin{tabular}{l c c c c c c}
\hline
 \multirow{2}{*}{Methods} & \multirow{2}{*}{Epochs} & \multirow{2}{*}{\# Regions} & \multicolumn{2}{c}{Medium} & \multicolumn{2}{c}{Hard} \\
 \cline{4-7}
 &  &  & $\mathcal R$-Oxf & $\mathcal R$-Par & $\mathcal R$-Oxf & $\mathcal R$-Par \\
 \hline
 WI & 500 & 1 & 8.3 & 38.1 & 1.4 & 9.9 \\
 GP~\cite{tolias2015particular} & 100 & 111 & 64.3 & 79.9 & 35.6 & 61.3 \\
 EB~\cite{zitnick2014edge} & 100 & 98 & 68.2 & 78.0 & 43.0 & 56.9 \\
 SS~\cite{uijlings2013selective} & 100 & 137 & 69.1 & 81.5 & 44.8 & 63.0 \\
 \hline
\end{tabular}
\end{center}
\end{table}

Table~\ref{tab:object-discovery} reports the result. As seen, compared with directly using the whole images (shown in the first row), using the obtained object regions (shown in the following three rows) clearly helps SSL to learn feature representations that can achieve better retrieval. This can be verified by the significant improvements on the mAP values for all the four retrieval cases. Also, to show the time spent in SSL learning process, this experiment reports the number of epochs required to reach convergence. It can be observed that these region-based SSL learning processes need only $100$ epochs to converge, while the whole image based one still fails after $500$ epochs. In addition, comparison within the three object proposal generators shows that selective search and edge boxes perform better than the grid-partition scheme.  This suggests that more accurate object region discovery may positively contribute to the quality of the feature representation learned by SSL. At the same time, considering that the grid-partition scheme is much simpler than the other two, its result has been very encouraging. In short, the above result verifies the necessity and advantage of utilising object regions for SSL in the proposed framework.

\subsubsection{Impacts of $\mathcal R$1M}
\begin{figure}
\centering
\includegraphics[width=3.2in]{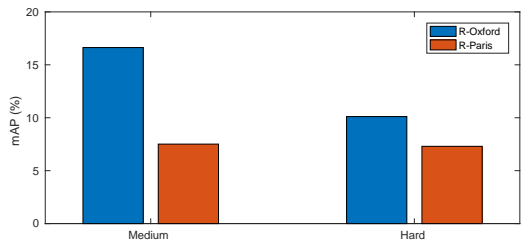}
\caption{Retrieval performance (in mAP) on our ``cross-distractors'' setting for $\mathcal R$-Oxford and $\mathcal R$-Paris.}
\label{fig:cross}
\end{figure}

To investigate the relationship between the $\mathcal R$1M and $\mathcal R$Oxford/$\mathcal R$Paris, we design an experiment that is evaluated by mAP to estimate their overlapping level on the setting we called ``cross-distractors''. In particular, we mixed the $\mathcal R$Oxford/$\mathcal R$Paris with $\mathcal R$1M together, and set each query-relevant image of one class from the former as query, the left images in this class as positives, all the remaining images in the former as `junk' images like~\cite{DBLP:conf/cvpr/PhilbinCISZ07}, and all the images from the distractor dataset as negatives. We conduct retrieval based on this setting and calculate the average precision among all the classes. By conducting retrieval under this setting, the mAP performance could reflect the level of intersection on the mixed dataset. In order to make a fair comparison, we use the feature extracted from ResNet101 pre-trained on ImageNet to represent the images. The performance is shown in Fig.~\ref{fig:cross}, as seen, on both ``Medium'' and ``Hard'' settings, the mAP values on $\mathcal R$Oxford are better than that of $\mathcal R$Paris. That means, more distractor images are distributed among the area of the query-relevant images in $\mathcal R$Paris than $\mathcal R$Oxford. This verifies our speculation.

\section{Conclusion}
This work investigates the potential of a novel framework for learning feature representation for instance image retrieval. Different from the existing retrieval methods that use either pre-trained deep models or external labeled datasets, this framework purely learns feature representation from a given retrieval dataset, with the goal to make it work best for retrieval on this specific dataset. The investigation shows surprisingly good performance of this framework, which can be comparable to the relevant state-of-the-art ones. This provides an attractive alternative to the existing approach for retrieval. Its advantages and limitations are also discussed. A lot of future work is worth exploring upon this new framework. Particularly, we focus on CNN based models to demonstrate the efficacy of the proposed framework in this work. Meanwhile, more advanced architectures such as ViT~\cite{DBLP:conf/iclr/DosovitskiyB0WZ21}, MLP~\cite{DBLP:conf/nips/TolstikhinHKBZU21}, and
hybrid variants have recently been developed. We will test our framework upon them in the future work.

\ifCLASSOPTIONcompsoc
  \section*{Acknowledgments} Zhongyan Zhang was supported by CSIRO Data61 PhD Scholarship and University of Wollongong International Postgraduate Tuition Award. Lei Wang and Zhongyan Zhang were supported by the Australian Research Council via the Discovery Project with grant number DP200101289. Jianjia Zhang was supported by the National Natural Science Foundation of China under Grant 62101611. This research was undertaken with the assistance of resources and services from the National Computational Infrastructure (NCI), which is supported by the Australian Government.
\else
  \section*{Acknowledgment}
\fi

\bibliography{egbib}
\bibliographystyle{IEEEtran}

\end{document}